\documentclass[letterpaper, 10 pt, journal, twoside, dvipsnames]{IEEEtran} 
\IEEEoverridecommandlockouts

\usepackage{etex}
\usepackage[T1]{fontenc}
\usepackage[utf8]{inputenc}
\usepackage{bbold}
\usepackage{cite}
\usepackage{soul}
\usepackage{graphicx}
\usepackage{epstopdf}
\usepackage{amssymb}
\usepackage[cmex10]{amsmath}
\usepackage{mathtools}
\usepackage{cases}

\usepackage{enumitem}
\usepackage{array}
\usepackage{multirow}
\usepackage{bigdelim}
\usepackage{mdwtab}
\usepackage{tikz}
\usepackage{physics}
\usepackage{color}
\usepackage{microtype}
\usepackage{eso-pic}

\makeatletter
\let\MYcaption\@makecaption
\makeatother
\usepackage[font=footnotesize]{subcaption}
\makeatletter
\let\@makecaption\MYcaption
\makeatother

\usepackage[usestackEOL]{stackengine}
\newcommand\fop[3][9pt]{\mathop{\ensurestackMath{\stackengine{#1}%
  {\displaystyle#2}{\scriptstyle#3}{U}{c}{F}{F}{L}}}\limits}

\DeclareMathOperator*{\argmin}{arg\,min}
\DeclareMathOperator*{\argmax}{arg\,max}

\newcommand\fmin[2][9pt]{\fop[#1]{\min}{#2}}

\graphicspath{{figs/}}

\usepackage{letltxmacro}
\LetLtxMacro{\originaleqref}{\eqref}
\renewcommand{\eqref}{Eq.~\originaleqref}

\usepackage{xcolor}
\usepackage{hyperref}

\hypersetup{
  linkcolor  = NavyBlue,
  citecolor  = NavyBlue,
  urlcolor   = NavyBlue,
  colorlinks = true,
}

\title{Intelligent Reference Curation for Visual Place Recognition via Bayesian Selective Fusion}%
\author{Timothy L.~Molloy$^{1}$, Tobias Fischer$^{2}$, Michael Milford$^{2}$ and Girish N.~Nair$^{1}$%
\thanks{Manuscript received: October 15, 2020; Accepted December 13, 2020.}%
\thanks{This paper was recommended for publication by Editor J.~Civera upon evaluation of the Associate Editor and Reviewers' comments. This work received funding from the Australian Government, via grant AUSMURIB000001 associated with ONR MURI grant N00014-19-1-2571. T.F.~and M.M.~acknowledge continued support from the Queensland University of Technology (QUT) through the Centre for Robotics. \textit{(Corresponding author: Timothy L.~Molloy)}}%
\thanks{$^{1}$Timothy L.~Molloy and Girish N.~Nair are with the Department of Electronic and Electrical Engineering, University of Melbourne, Parkville, VIC 3010, Australia. {\texttt{\{tim.molloy, gnair\}@unimelb.edu.au}}.}
\thanks{$^{2}$Tobias Fischer and Michael Milford are with the QUT Centre for Robotics, Queensland University of Technology, Brisbane, QLD 4000, Australia. {\texttt{\{tobias.fischer, michael.milford\}@qut.edu.au}}.}
\thanks{Digital Object Identifier (DOI): \href{http://doi.org/10.1109/LRA.2020.3047791}{10.1109/LRA.2020.3047791}}%
}

\begin{document}
\bstctlcite{IEEEexample:BSTcontrol}

\AddToShipoutPicture*{%
     \AtTextUpperLeft{%
         \put(-3.5,10){
           \begin{minipage}{\textwidth}
              \scriptsize
              Preprint version; final version available at \url{http://doi.org/10.1109/LRA.2020.3047791}
           \end{minipage}}%
     }%
}

\maketitle

\begin{abstract}
\boldmath
A key challenge in visual place recognition (VPR) is recognizing places despite drastic visual appearance changes due to factors such as time of day, season, weather or lighting conditions.
Numerous approaches based on deep-learnt image descriptors, sequence matching, domain translation, and probabilistic localization have had success in addressing this challenge, but most rely on the availability of carefully curated representative reference images of the possible places.
In this paper, we propose a novel approach, dubbed Bayesian Selective Fusion, for actively selecting and fusing informative reference images to determine the best place match for a given query image.
The selective element of our approach avoids the counterproductive fusion of every reference image and enables the dynamic selection of informative reference images in environments with changing visual conditions (such as indoors with flickering lights, outdoors during sunshowers or over the day-night cycle).
The probabilistic element of our approach provides a means of fusing multiple reference images that accounts for their varying uncertainty via a novel training-free likelihood function for VPR.
On difficult query images from two benchmark datasets, we demonstrate that our approach matches and exceeds the performance of several alternative fusion approaches along with state-of-the-art techniques that are provided with prior (unfair) knowledge of the best reference images. 
Our approach is well suited for long-term robot autonomy where dynamic visual environments are commonplace since it is training-free, descriptor-agnostic, and complements existing techniques such as sequence matching.

\end{abstract}

\IEEEpeerreviewmaketitle

\section{Introduction}
\IEEEPARstart{V}{isual} place recognition (VPR) is the problem of determining the place at which a given query image was captured and is a key enabling technology for mobile robot localization.
The overwhelming majority of VPR approaches involve comparing query images with reference images previously captured at each candidate place in a map or database of the environment \cite{Lowry2016}.
Research on VPR over several decades \cite{Lowry2016} has therefore explored techniques for robust image comparison including deep-learnt image descriptors \cite{NetVLAD, DenseVLAD, DeltaDescriptors}, sequence matching \cite{Milford2012, Naseer2018, vysotska2019effective}, and multiprocess fusion \cite{hausler2019multi}.

\begin{figure}[t!]
    \centering
    \includegraphics[width=0.8\columnwidth]{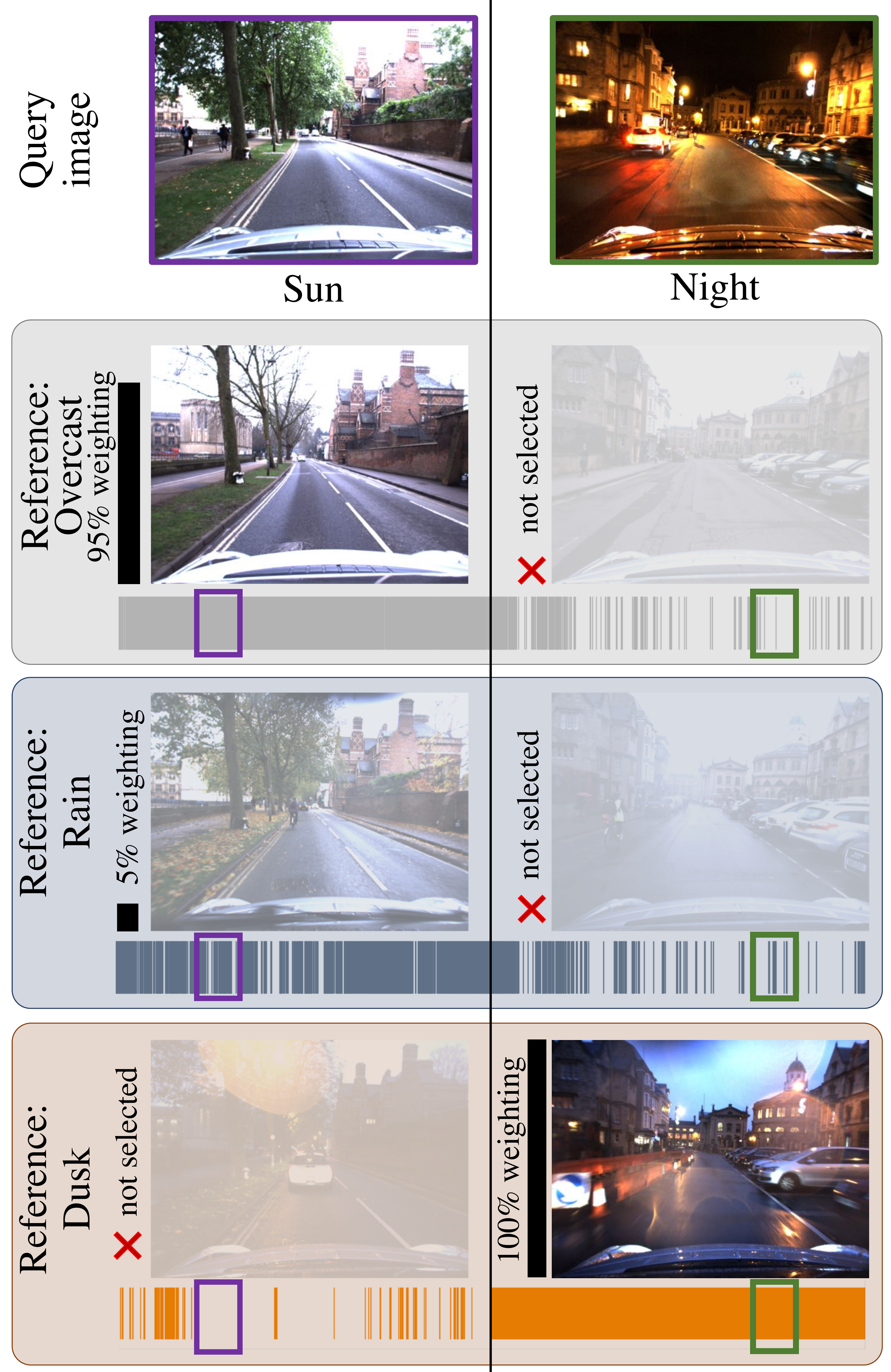}
    \vspace*{-0.2cm}
    \caption{Given a query image, Bayesian Selective Fusion selects and fuses informative reference images to find the best place match. Left: Given sunny query images, our approach selects images from the overcast and rain reference sets (frequently discarding the dusk images), and fuses them according to their likelihood (higher transparency indicates lower likelihood; see also black bars). Right: Given night query images, our approach selects images from the dusk reference set (frequently discarding the overcast and rain images) since they are perceptually most similar to night images. Note however that when dusk images offer limited visual information (e.g., due to sun glare), additional images from the rain and overcast references are selected and fused.
    }
    \vspace*{-0.25cm}
    \label{fig:my_label}
\end{figure}

The need for representative place images in VPR has also led to the development of techniques such as voting \cite{kovsecka2005global}, experience maps \cite{linegar2015work, Doan2019ICCV}, memory compression \cite{rosman2014coresets} and domain translation \cite{Anoosheh2019} to curate references images.
The majority of these approaches are learning-based, and determine the utility of reference images during training to select which to store or learn from.
Determining the utility of reference images is however inherently challenging during training since ultimately it depends on factors affecting the similarity of the place appearance at query time. 
Surprisingly few approaches have sought to address this challenge since the rise of deep-learnt image descriptors for VPR in, e.g.,~\cite{NetVLAD,chen2017deep,khaliq2019holistic,chen2017only,schubert2020unsupervised,vysotska2019effective}.

Most recently, \cite{Fischer2020} has achieved state-of-the-art performance in a challenging VPR scenario with event-cameras using a training-free fusion approach for deep-learnt image descriptors.
Although achieving state-of-the-art performance, the approach of \cite{Fischer2020} fuses all reference images at a given place.
This complete fusion can be unnecessary and even counterproductive, especially in dynamic visual environments where query images can originate from strikingly different conditions --- for example, indoors with lights flickering or outdoors with cloud and rain showers occluding the sun.
In this paper, we propose a novel approach to intelligently fuse informative reference images to avoid unnecessary or counterproductive fusion. 

The key contributions of this paper are:
\begin{enumerate}
    \item
    A novel Bayesian Selective Fusion approach for \emph{single-image} VPR that intelligently selects informative reference images and fuses these using Bayesian data fusion.
    \item
    A novel training-free likelihood for probabilistic VPR that is agnostic to the underlying descriptors.
    \item
    Demonstration of the state-of-the-art performance of our approach compared to the state-of-the-art approach of \cite{Fischer2020} and a suite of other fusion and non-fusion approaches.
\end{enumerate}
Our approach complements existing VPR techniques (e.g.~sequence matching) by providing a dynamic, training-free, descriptor-agnostic, and probabilistic means of exploiting the information from multiple (same-place) reference images.

This paper is structured as follows:
Section \ref{sec:related} presents related work; Section \ref{sec:approach} describes our proposed Bayesian Selective Fusion approach; Section \ref{sec:setup} reports our experimental setup; Section \ref{sec:results} presents the results of our experiments; and, Section \ref{sec:conclusion} offers conclusions and suggestions for future work.

\section{Related Work}
\label{sec:related}
We first review probabilistic approaches to VPR and approaches that adapt the visual conditions of images before place matching. We then provide an overview of how previous approaches fuse multiple sets of reference images.

\subsection{Probabilistic and Image Adaption Approaches}
\label{subsec:rel:bayesian}
Arguably the most prominent probabilistic approach to recognizing places is FAB-MAP~\cite{cummins2008fab}. One of FAB-MAP's main innovations was to explicitly account for perceptual aliasing, such that highly similar but indistinct observations receive a low probability as being the same place; a generative model of bag-of-words observations implemented this. FAB-MAP has been extended in multiple ways, including the incorporation of odometry information and image sequences \cite{maddern12catslam, maddern12catgraph}.

While FAB-MAP relies on local image features, Lowry et al.~\cite{lowry2014towards} present a probabilistic model for whole-image descriptors, which leads to greater robustness in environments with large perceptual changes. Ramos et al.~\cite{ramos2012bayesian} propose a Bayesian framework for place recognition that leverages \mbox{5-9} training images per place and can then recognize a place from previously unseen viewpoints. Recently, Oyebode et al.~\cite{oyebode2019sample} have proposed a Bayesian framework that leverages pre-trained object detectors to recognize indoor places based purely on object categories, which is related to the concept of visual place \emph{categorization}~\cite{wu2009visual}, i.e.~predicting the semantic category of a place.

Dubios et al.~\cite{Dubois2011} use Bayesian filtering to model the dependency between sequences of images. Similar probabilistic temporal sequence modeling has also been investigated in \cite{Naseer2018} and in \cite{doan2019visual} using a Monte Carlo-based algorithm that achieves state-of-the-art performance on several datasets.
In contrast to these methods that use sequences to temporally fuse place information, our VPR approach operates on single query images and fuses multiple reference images. 

Our work is also related to approaches that preprocess query images to match conditions in reference images (or \emph{vice versa}).
For example, Pepperell et al.~\cite{pepperell2016routed} and recent domain translation methods (cf.~\cite{Anoosheh2019}) cast query images into the same visual conditions as the reference images via sky removal or night-to-day image translations (see also \cite{saurer2016image}, \cite[Section VII-A]{Lowry2016} and references therein). 
However, these approaches still raise the question of which reference images and translations are best to use for a given query image. For example, night-to-day translation implicitly assumes that query images are known to have been captured at night, and is unnecessary and could degrade performance on daytime query images.

\subsection{Fusion of Reference Images}
\label{subsec:rel:fusion}

Kosecka et al.~\cite{kovsecka2005global} described indoor places with a set of \emph{representative views} and then found the most likely reference image using a maximum voting scheme. Carlevaris-Bianco and Eustice~\cite{carlevaris2012learning} proposed an extension to that approach by learning temporal observability relationships between the representative views. In a similar vein, the approach by Johns and Yang~\cite{johns2013feature} learns feature co-occurrence statistics, such that each place is described using a set of co-occurring features. This allows for reliable matching of images that were captured at different times of the day.

When robots are deployed for prolonged periods, it is common to update the environment model over time. Biber and Duckett represent the environment using multiple timescales simultaneously, whereby older memories are faded at a rate that depends on the particular timescale being used~\cite{biber2009experimental}. For each place, the timescale with the highest log-likelihood is chosen. A more sophisticated approach in robot navigation is to store only persistent configurations in a long-term memory, so that dynamic objects do not become part of the map~\cite{morris2014multiple}. Another approach is the use of \emph{coresets}~\cite{rosman2014coresets} to drastically compress the images that represent a place. These coresets have subsequently been used to detect loop closures in real-time~\cite{volkov2015coresets}.

Impressive results have been achieved using so-called experience maps~\cite{linegar2015work}. The accumulation of multiple experiences for the same place allows for both adaptation to changing appearance due to lighting and weather conditions as well as structural changes. At localization time, the robot can then predict the most appropriate experience using a probabilistic framework. Selecting the most appropriate experience results not only in reduced computational time but also in significantly reduced failure rates. A similar method by Doan et al.~\cite{Doan2019ICCV} accumulates additional experiences for the same place while avoiding an unbounded growth of computation and storage. Vysotska and Stachniss~\cite{vysotska2019effective} match places in a data association graph that can contain images of multiple reference traverses.

Further references to methods that contain multiple representations of the environment can be found in the survey on visual place recognition by Lowry et al.~\cite{Lowry2016} in Section VII-B.2. While our method aims to leverage the complementary nature of multiple reference traverses, the multiprocess fusion approach by Hausler et al.~\cite{hausler2019multi} fuses predictions of multiple complementary image processing methods applied to the same input image. This also relates to~\cite{lowry2016supervised}, where appearance changes between the query and reference traversals are removed to some degree by using as few as 100 training samples. Contrary to~\cite{lowry2016supervised}, our method is training free. Our method is also related to~\cite{schubert2020unsupervised}, where in-sequence condition changes are investigated; however, only a single reference set is used.

\section{Approach}
\label{sec:approach}

In this section, we present our novel Bayesian Selective Fusion approach.
We first revisit the formulation and solution of \emph{single-image} VPR based on descriptor distances, including the state-of-the-art fusion approach of \cite{Fischer2020}.
We exploit this formulation and the properties of descriptor distances to propose a novel strategy for selecting reference images and a new Bayesian method for fusing them, along with a novel likelihood function for probabilistic VPR.

\subsection{Problem Formulation and the Minimum-Value Principle}
\label{subsec:VPRformulation}
Let $X$ be an unknown place at which a robot captures a \emph{query} image $I_X$ and extracts a corresponding image descriptor vector $z_X \in \mathbb{R}^{N_z}$ with dimension $N_z$ (e.g., using NetVLAD~\cite{NetVLAD}).
Let $X$ belong to the set $\mathcal{X} = \{1, \ldots, N\}$ of possible places.
The robot has access to \emph{reference} images of the places in $\mathcal{X}$ captured during previous visits under different visual appearance conditions.
These reference images are stored in a total of $M$ \emph{reference sets}, with individual reference sets indexed by the scalar $u \in \{1, \ldots, M\} \triangleq \mathcal{U}$.
Each reference set $u \in \mathcal{U}$ contains (at most) one image from each place in $\mathcal{X}$ (together with associated image descriptor).
Let $I_i^u$ and $z_i^u \in \mathbb{R}^{N_z}$ denote the image and corresponding descriptor from reference set $u \in \mathcal{U}$ corresponding to place $i\in \mathcal{X}$.
Given a reference set $u \in \mathcal{U}$, the robot can compute non-negative distances $d : \mathbb{R}^{N_z} \times \mathbb{R}^{N_z} \mapsto \mathbb{R}_{\geq0}$ between the descriptor $z_X$ of the query image $I_X$ and the descriptors $z_i^u$ from the reference set $u$ for each place $i$ in $\mathcal{X}$.
We collect the distances corresponding to reference set $u$ in the vector
\begin{align}
    D^u
    \triangleq
    [
    d(z_X, z_1^{u}) \quad d(z_X, z_2^{u}) \quad \cdots \quad d(z_X, z_N^{u})
    ].
\end{align}
We shall use $D_i^u$ to denote the $i$th component of $D^u$, i.e. $D_i^u \equiv d(z_X, z_i^u)$, and without loss of generality, we consider the Euclidean distance for $d(z_X, z_i^u)$ (other distance measures could also be used).
The VPR problem is to infer the place $X$ using the vectors $D^u$ from some or all of the reference sets $\mathcal{U}$.

The vast majority of recent VPR approaches rely on the \emph{minimum-value principle} (or best-match-strategy) \cite{Naseer2018, lowry2014towards}.
The principle asserts that the best place match $\hat{X}$ corresponds to the minimum descriptor distance in each reference set, i.e.,
\begin{align}
\label{eq:mvp}
 X \approx \hat{X}
 = \argmin_{i \in \mathcal{X}} D_i^u
\end{align}
for any $u \in \mathcal{U}$.
The \emph{minimum ensemble distance} approach of \cite{Fischer2020} is a recently proposed state-of-the-art method based on the minimum-value principle that fuses information from multiple reference sets.
In this minimum ensemble distance approach, place matches $\hat{X}$ are found by minimizing the average (equivalently the total) of the descriptor distance vectors $D^u$ over all reference sets, namely,
\begin{align}
    \label{eq:MED}
    \hat{X}
    = \argmin_{i \in \mathcal{X}} \sum_{u = 1}^M D_i^u.
\end{align}

Clearly, VPR approaches based on the minimum-value principle perform poorest when the place that minimizes the descriptor distances is not the true place --- a situation which for example might occur when the query and reference images are captured under different appearance conditions.
Whilst their robustness can be improved by fusing reference sets captured under a variety of appearance conditions using addition or averaging (as in \eqref{eq:MED}), these operations implicitly assume that all reference sets provide equally useful information.

We hypothesize that reference sets will have different utility for different query images and that this utility will be inherently uncertain.
Therefore, we propose a novel Bayesian selective fusion approach that: 1) selects informative reference sets based on the minimum-value principle; and, 2) fuses the selected reference sets using Bayesian fusion to handle the uncertainty of the minimum-value principle holding.

\subsection{Proposed Reference Set Selection}
\label{subsec:refsetselection}
Given the distance vectors $D^u$ from all reference sets $u \in \mathcal{U}$, we select a variable subset of reference sets, defined as
\begin{align}
    \label{eq:reference_selection}
    \mathcal{S}
    &= \left\{ u \in \mathcal{U} : \dfrac{\min_{i \in \mathcal{X}} D_i^{u} - \min_{i \in \mathcal{X}} D_i^{u^*}}{\min_{i \in \mathcal{X}} D_i^{u^*}} \leq \gamma \right\} \subset \mathcal{U},
\end{align}
with which to compute a place decision.
The collection $\mathcal{S}$ is formed by first selecting the ``best'' reference set under the minimum-value principle, that is, the reference set $u^*$ that contains the minimum distance such that
\begin{align}
    \label{eq:best_reference}
    u^*
    &\triangleq \argmin_{u \in \mathcal{U}} \fmin{i \in \mathcal{X}} D_i^u.
\end{align}
All reference sets $u \in \mathcal{U}$ with minimum descriptor distances $\min_{i \in \mathcal{X}} D_i^u$ within a fraction $\gamma > 0$ of the minimum descriptors distance $\min_{i \in \mathcal{X}} D_i^{u^*}$ of the reference set $u^*$ are also added to $\mathcal{S}$.
Our reference set selection approach resembles outlier rejection with the parameter $\gamma$ controlling the rejection of reference sets from $\mathcal{U}$ with minimum descriptor distances that are outliers compared to the ``best'' reference set $u^*$.

\subsection{Proposed Bayesian Fusion}
Given the descriptor distance vectors $\mathcal{D}^{\mathcal{S}} \triangleq \{D^u : u \in \mathcal{S}\}$ from the selected reference sets $\mathcal{S}$, the Bayesian belief $P(X | \mathcal{D}^{\mathcal{S}})$ over the places in $\mathcal{X}$ is given by Bayes' rule:
\begin{align}
    \label{eq:bayes}
    P(X = i | \mathcal{D}^{\mathcal{S}})
    &= \dfrac{P(\mathcal{D}^{\mathcal{S}} | X = i) P(X = i)}{\sum_{j = 1}^N P(\mathcal{D}^{\mathcal{S}} | X = j) P(X = j)}
\end{align}
for $i \in \mathcal{X}$.
Any prior knowledge that the robot has about its location (arising for example from its motion model or previous place matches) is incorporated here through the prior probability distribution $P(X)$. 
In the absence of any prior knowledge, this is taken as uniform (i.e. $P(X = i) = N^{-1}$ for all $i \in \mathcal{X}$).
The likelihood $P(\mathcal{D}^{\mathcal{S}} | X)$ enables our novel fusion of the place information from the selected references $\mathcal{S}$.

To make the novel fusion operation in \eqref{eq:bayes} explicit (and scalable), note that the descriptor distances $D_i^u$ are solely functions of the reference image descriptors $z_i^u$ after $X$ (and hence $I_X$ and $z_X$) is given.
The vectors $D^u$ from different reference sets are thus conditionally independent given $X$, and the likelihood $P(\mathcal{D}^{\mathcal{S}} | X)$ simplifies to the product:
\begin{align}
    \label{eq:product_form}
    P(\mathcal{D}^{\mathcal{S}} | X = i)
    &= \prod_{u \in \mathcal{S}} P(D^u | X = i)
\end{align}
for $i \in \mathcal{X}$ where $P(D^u | X)$ are the likelihoods for the single (individual) reference sets $u \in \mathcal{S}$.
Constructing the \emph{single-reference} likelihoods $P(D^u | X)$ is easier and more scalable than constructing the joint likelihood $P(\mathcal{D}^{\mathcal{S}} | X)$ as we shall discuss in the following subsection.
Here, we note that since the likelihood $P(\mathcal{D}^{\mathcal{S}} | X)$ in \eqref{eq:product_form} has a product form and the denominator in \eqref{eq:bayes} serves only to normalize, we may rewrite \eqref{eq:bayes} as simply:
\begin{align}
    \label{eq:unnormalized_bayes}
    P(X = i | \mathcal{D}^{\mathcal{S}})
    &\propto \prod_{u \in \mathcal{S}} P(D^u | X = i) P(X = i)
\end{align}
for $i \in \mathcal{X}$.
The robot's place recognition decision with our approach, denoted $\hat{X}$, is then the place with the maximum (unnormalized) Bayesian belief, namely,
\begin{align}
    \label{eq:approach}
    \hat{X}
    &= \argmax_{i \in \mathcal{X}} \prod_{u \in \mathcal{S}} P(D^u | X = i) P(X = i).
\end{align}
No place recognition decision is made if the belief fails to exceed a threshold $h > 0$, which balances recognizing the wrong place (small $h$) with missing the true place (large $h$).
We next describe a novel efficient construction of the \emph{single-reference} likelihoods $P(D^u | X)$ used in \eqref{eq:approach}.

\subsection{Proposed Training-Free Single-Reference Likelihoods}
To construct the \emph{single-reference} likelihoods $P(D^u | X)$, note that by definition they are the joint likelihoods of the distances, i.e., $P(D^u | X) \equiv P(D_1^u, \ldots, D_N^u | X)$.
Note also that in order for the distances $D_i^u$ to all be equally useful, their (marginal) distribution should depend only on whether the places $X$ \mbox{and $i$} match or not, and not on the specific places $X$ and $i$.
Hence, we model the distances $D_i^u$ as conditionally independent given $X$ and distributed according to a (marginal) distribution $P(D_i^u | X \neq i)$ when the place $X$ does not match $i$ and a different (marginal) distribution $P(D_i^u | X = i)$ when the place $X$ matches $i$.
The (joint) likelihood of any place $i$ after observing the vector $D^u$ from reference set $u$ is thus the product of the (marginal) likelihoods after observing the individual distances, namely,
\begin{equation*}
    P(D^u | X = i)
    = P(D_i^u | X = i) \prod_{\substack{j = 1, j \neq i}}^N P(D_j^u | X \neq j)
\end{equation*} 
for $i \in \mathcal{X}$.
Equivalently, we may write this likelihood as
\begin{equation}
    \label{eq:obsLikelihoods}
    P(D^u | X = i)
    \propto
     \dfrac{P(D_i^u | X = i)}{P(D_i^u | X \neq i)}
\end{equation}
for $i \in \mathcal{X}$. Here, the proportionality constant $C^u \triangleq \prod_{j = 1}^N P(D_j^u | X \neq j)$ is constant for all $i \in \mathcal{X}$, which allows us to simply ignore it in our approach (cf.~\eqref{eq:approach}).

Place-match and non-place-match likelihoods, $P(D_i^u | X = i)$ and $P(D_i^u | X \neq i)$, have previously been constructed via training for (whole-image) descriptors (cf.~\cite{lowry2014towards, Lowry2016}).
These approaches can be used directly to evaluate \eqref{eq:obsLikelihoods}, however we propose an alternative training-free approach based on a probabilistic encoding of the minimum-value principle (\eqref{eq:mvp}).
For a given reference set $u$, we model the place-match likelihood $P(D_i^u | X = i)$ as being proportional to the number of places with descriptor distances larger than $D_i^u$ in $D^u$, i.e.,
\begin{align}
    \label{eq:placeMatch}
    P(D_i^u | X = i)
    \propto \sum_{j = 1}^N \mathbb{1}\{ D_i^u \leq D_j^u  \}
\end{align}
where the indicator function $\mathbb{1}\{ D_i^u \leq D_j^u  \}$ takes a value of $1$ when $D_i^u \leq D_j^u$ and is zero otherwise.
Again, the proportionally constant is independent of $i$, and can be ignored.

We model the non-place-match likelihood $P(D_i^u | X \neq i)$ by recalling that the distribution of descriptor distances is independent of the place, and hence most elements in the vector $D^u$ are realizations of the descriptor distance with $X \neq i$.
Due to the computational efficiency of computing the mean $\mu$ and variance $\sigma^2$ of the elements in $D^u$, we simply model the non-place-match likelihood as the Gaussian $P(D_i^u | X \neq i) = \mathcal{N} ( D_i^u; \mu, \sigma^2 )$ (accepting that some bias will be incurred due to one element $i$ of $D^u$ corresponding to $X = i$).

\subsection{Summary of Bayesian Selective Fusion}
In summary, our proposed approach first selects reference sets according to \eqref{eq:reference_selection}, then computes the single-reference likelihoods given by \eqref{eq:obsLikelihoods} using \eqref{eq:placeMatch} and our Gaussian construction of non-place-match likelihood, and finally declares a place recognition decision via the maximization in \eqref{eq:approach}.

\section{Experimental Setup}
\label{sec:setup}

In this section, we describe the datasets, methods, metrics, and parameters with which we evaluate our approach.

\subsection{Datasets}
We use two, widely used, established benchmark datasets.

The \textbf{Nordland} dataset~\cite{sunderhauf2013we} consists of four recordings of a train traversing a 728km long route in Norway captured in spring, summer, fall and winter. We extract one image per second from a one hour and twenty minute long section of these videos (20:00 to 1:40:00) and, as typically done in the literature, manually remove tunnels and sections where the train is stationary or traveling at speeds below 15km/h (based on the available GPS information). This resulted in a total of 3000 frames per season. As frame $i$ in one of those videos corresponds exactly to frame $i$ in the other videos, we find ground-truth correspondences by matching the query traverse frame number with the reference traverse number and allow a ground-truth tolerance of $\pm2$ frames. The Nordland dataset has been widely used in the literature, e.g.~in~\cite{hausler2019multi,DeltaDescriptors,lowry2016supervised}.

The \textbf{Oxford RobotCar} dataset~\cite{maddern20171} contains over 100 traversals of a consistent route through Oxford, captured at different times of the day and in varying weather conditions. Recently, centimeter-accurate ground-truth annotations (based on post-processed GPS, IMU and GNSS base station recordings) were made available for a subset of these traversals~\cite{RCDRTKArXiv}. We selected five traversals of the same route\footnote{Dusk: 2014-11-21-16-07-03, Night: 2014-12-16-18-44-24, Overcast: 2015-05-19-14-06-38, Sun: 2015-08-12-15-04-18, Rain: 2015-10-29-12-18-17} and subsampled them such that places $i$ and $i+1$ have a regular spatial separation of one meter between them, which resulted in 3350 frames. We used images recorded using the front left stereo camera. Similarly to the Nordland dataset, the Oxford RobotCar dataset is widely used in the literature, e.g.~see~\cite{hausler2019multi,DeltaDescriptors}. As in~\cite{DeltaDescriptors}, we use a ground-truth tolerance of $\pm10$ meters.

We computed NetVLAD \cite{NetVLAD} descriptors for the images extracted from the datasets.
Following \cite{NetVLAD} and as has become standard in subsequent studies (cf.~\cite{Fischer2020,DeltaDescriptors}), the descriptors were reduced to 4096 components via PCA.
We use these NetVLAD descriptors for $z_X$ and $z_i^u$ in all experiments except for one comparative study (Section \ref{subsec:study_dense}), for which we also compute 4096-d DenseVLAD \cite{DenseVLAD} descriptors.

\subsection{Baseline, Fusion, and Single-Reference Methods}
We use the state-of-the-art minimum ensemble distance approach of \cite{Fischer2020} (see also \eqref{eq:MED}) as the \emph{Baseline Fusion} approach for comparison with our proposed \emph{Bayesian Selective Fusion} approach.
The reference sets used by these fusion methods constitute the remainder of the traversals on the relevant dataset after removing the query traversal (i.e., if the query is Nordland Summer, then the reference sets are Winter, Fall and Spring).
We also implement standard \emph{single-reference} NetVLAD (cf.~\eqref{eq:mvp}) and versions of our Bayesian approach with single reference sets (i.e. $M = 1$).
These single-reference methods are labeled according to their reference set (e.g., given only a Nordland spring reference, NetVLAD is termed \emph{NetVLAD Spring} and the single-reference Bayesian approach is termed \emph{Bayesian Spring} or simply \emph{Spring}).

\subsection{Performance Metrics}
We report VPR performance using Precision-Recall (PR) curves (with the threshold $h$ swept for our proposed method).
Precision is defined as the ratio of correct place matches to the total number of place matches whilst recall is defined as the ratio of correct place matches to the total number of possible true matches.
Our datasets have a true match for every query image.
In some experiments, we also report the area under the PR curve (AUC) as a summary statistic.
As in \cite{chen2017deep,khaliq2019holistic,chen2017only}, the AUC serves as a proxy for recall at 100\% precision (which is of direct concern for SLAM) for comparing single-image VPR methods since many offer no recall at 100\% precision.

\subsection{Parameter: Fraction for Selecting Reference Sets\texorpdfstring{ $\gamma$}{}}

Our approach relies on a single parameter, $\gamma$ (see \eqref{eq:reference_selection}).
In our experience, values of $\gamma \in [0.04,0.1]$ provide reasonable performance on the Nordland and Oxford datasets.
To illustrate this insensitivity, we use a fixed $\gamma = 0.04$ in all experiments, noting that improvements could be obtained by tuning $\gamma$.

\section{Experimental Results}
\label{sec:results}
In this section, we evaluate our approach and its reference selection and Bayesian aspects across six experiments, including comparisons and extensions to state-of-the-art techniques.
In Section~\ref{subsec:study_1}, we first examine the performance offered by our new training-free likelihood for Bayesian VPR in comparison to NetVLAD.
In Sections~\ref{subsec:study_2} and \ref{subsec:study_3}, we compare our Bayesian Selective Fusion approach with the state-of-the-art fusion approach of \cite{Fischer2020} and NetVLAD (with NetVLAD provided advantageously with the best performing reference set). 
In Section~\ref{subsec:study_4} we showcase our approach in a scenario with extreme appearance changes.
Finally, in Sections~\ref{subsec:study_seq} and~\ref{subsec:study_dense} we demonstrate that our approach can also exploit sequence matching (i.e.~SeqSLAM~\cite{Milford2012}) and alternative image descriptors (i.e.~DenseVLAD~\cite{DenseVLAD}).

\subsection{Single-Reference Method Comparison}
\label{subsec:study_1}
\begin{figure}[t!]
    \vspace{0.2cm} %
    \centering
    \includegraphics[width = 0.9\columnwidth]{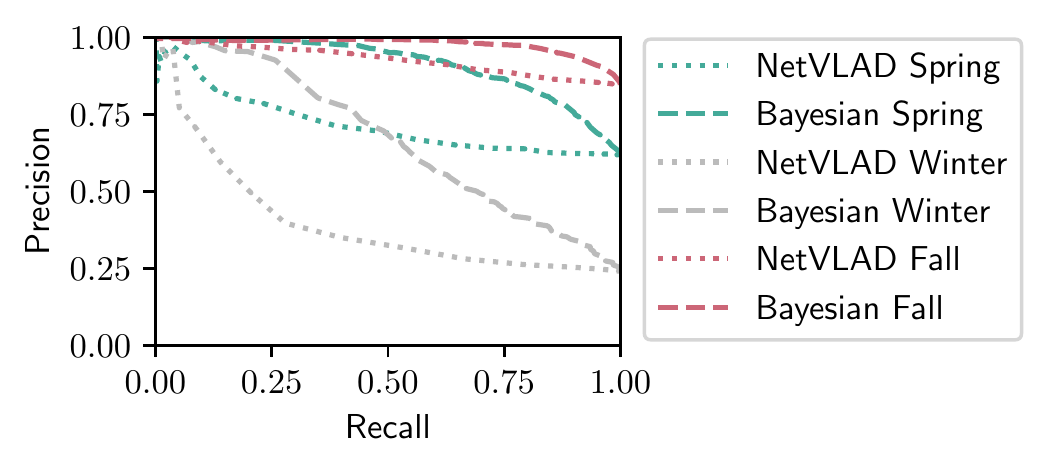}\vspace*{-0.3cm}
    \caption{Precision-recall of single-reference (no fusion) Baseline and proposed Bayesian methods on a Nordland Summer query traversal.}
    \label{fig:study_1}
\end{figure}

Fig.~\ref{fig:study_1} shows that Bayesian approaches with our new training-free \emph{single-reference} likelihood outperforms NetVLAD given the same reference set for query images drawn from the Nordland summer traversal.
The Bayesian approach offers a $5.5\%$ (relative) improvement in AUC over NetVLAD when both perform their best (Fall reference) and a $70\%$ improvement when both perform their worst (Winter reference).
Fig.~\ref{fig:study_1} motivates the use of (selective) fusion since the performance of the methods is strongly dependent on the reference images they are provided.
We thus now focus on evaluating the performance of our Bayesian Selective Fusion approach.

\subsection{Bayesian Selective Fusion vs Baseline Fusion}
\label{subsec:study_2}
\begin{figure}[t!]
    \vspace{0.2cm} %
    \centering
    \includegraphics[width=\columnwidth]{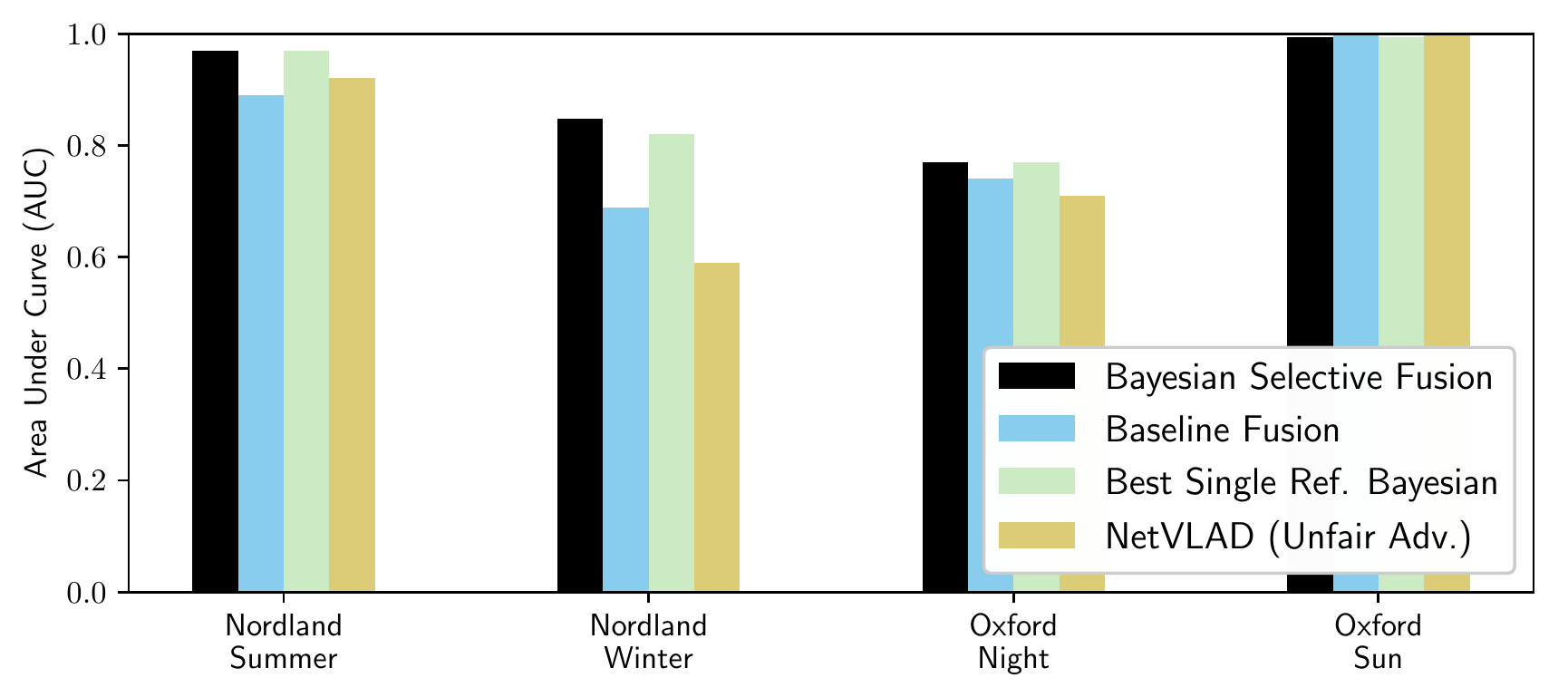}
    \caption{Area Under Precision-Recall Curves of Fig.~\ref{fig:study_2} for query images from four traverses from Nordland and Oxford RobotCar datasets.}
    \label{fig:study_2_auc}
    \vspace{-0.1cm}
\end{figure}

\begin{figure}[t!]
    \captionsetup[subfigure]{aboveskip=-1.5pt,belowskip=0.5pt,labelformat=simple}
    \renewcommand*{\thesubfigure}{\quad(\alph{subfigure})}
    \centering
    \includegraphics[width = 0.7\columnwidth]{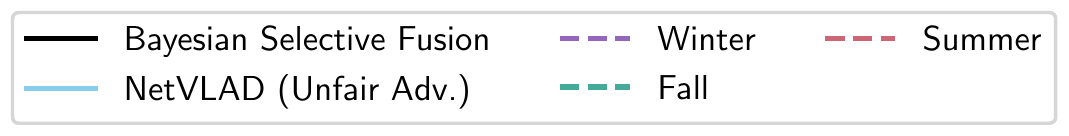}\\
    \begin{subfigure}{0.49\columnwidth}
         \centering
         \includegraphics[width=\textwidth]{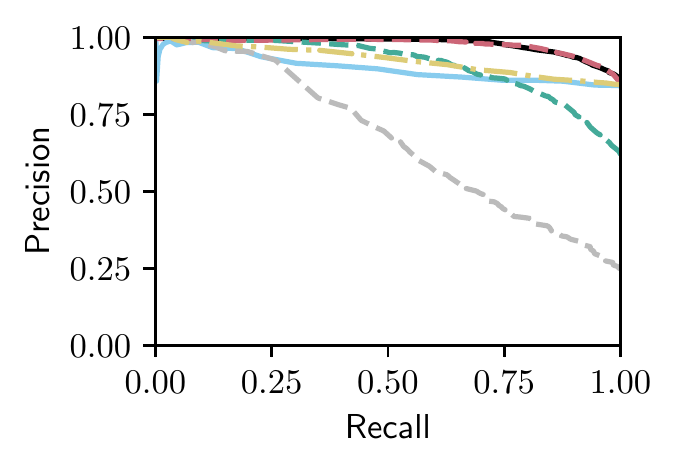}
         \caption{Query: Nordland Summer}
     \end{subfigure}
     \hfill
     \begin{subfigure}{0.49\columnwidth}
         \centering
         \includegraphics[width=\textwidth]{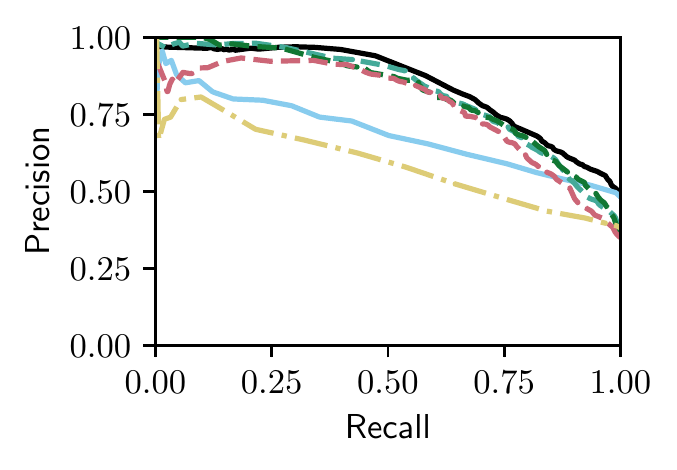}
         \caption{Query: Nordland Winter}
     \end{subfigure}\\
     \includegraphics[width = 0.99\columnwidth]{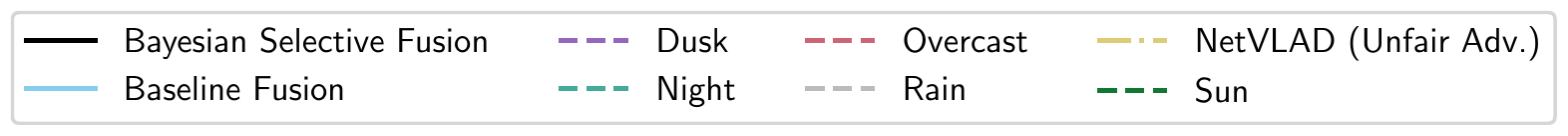}\\
     \begin{subfigure}{0.49\columnwidth}
         \centering
         \includegraphics[width=\textwidth]{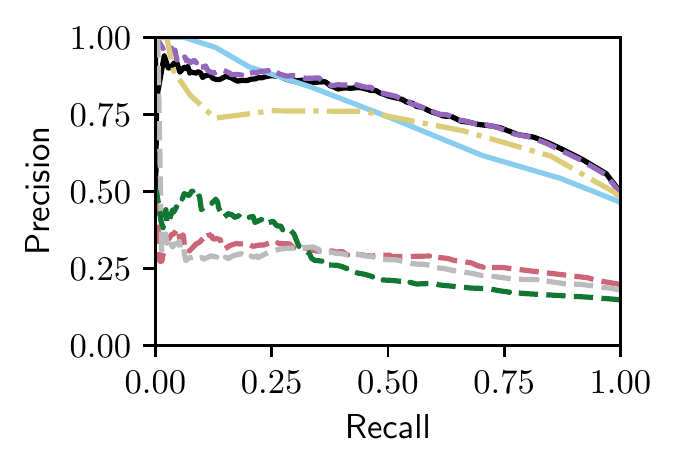}
         \caption{Query: Oxford Night}
     \end{subfigure}
     \hfill
     \begin{subfigure}{0.49\columnwidth}
         \centering
         \includegraphics[width=\textwidth]{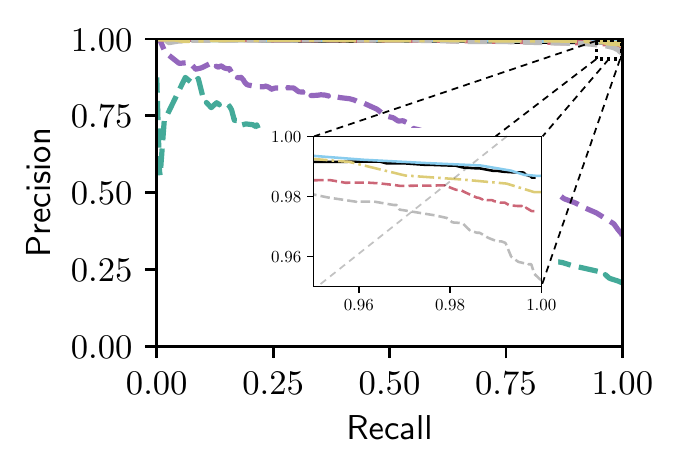}
         \caption{Query: Oxford Sun}
     \end{subfigure}
    \caption{Precision-recall comparison of Bayesian Selective Fusion, Baseline Fusion, and single-reference Bayesian methods on query images from four traverses from Nordland and Oxford RobotCar datasets.}
    \label{fig:study_2}
    \vspace{-0.2cm}
\end{figure}

We next compare our Bayesian Selective Fusion approach with the state-of-the-art fusion approach of \cite{Fischer2020} (i.e., Baseline Fusion) and NetVLAD (unfairly) given the best performing reference on each traversal.
Figs.~\ref{fig:study_2_auc}~and~\ref{fig:study_2} show that our approach outperforms the Baseline approach by a large margin on Nordland Summer and Winter query traversals with AUCs of $0.97$ and $0.84$ for our approach compared to $0.89$ and $0.68$ for the baseline method ($9\%$ and $24\%$ increase, respectively).
Our approach performs slightly worse than the Baseline Fusion approach on the easier Oxford Sun traversal (with an AUC of $0.995$ compared to $0.997$) but outperforms it on the more difficult Oxford Night traversal, with an AUC of $0.77$ compared to $0.74$ ($4\%$ increase).

\begin{figure}[t!]
    \centering
    \includegraphics[width = 0.75\columnwidth]{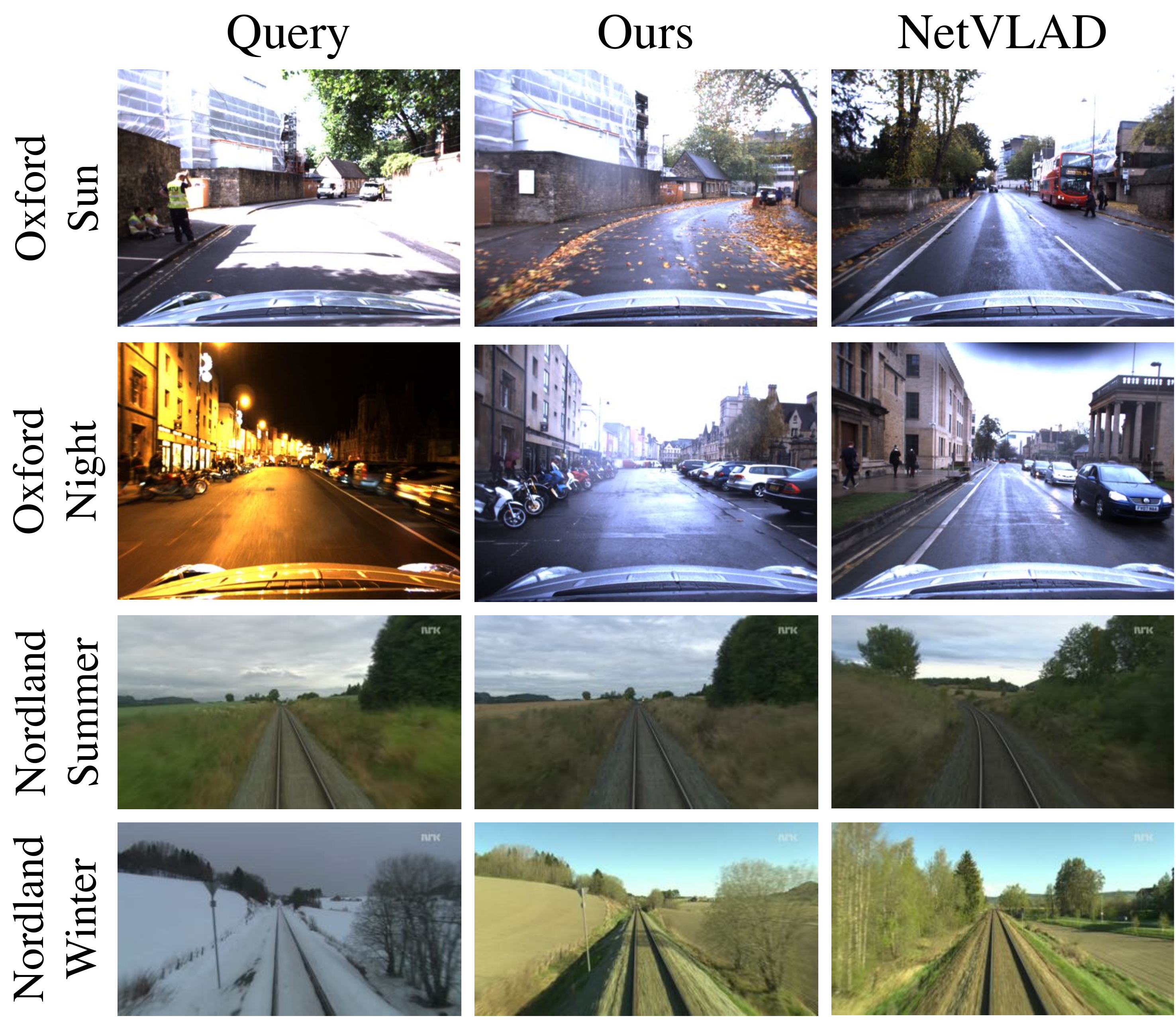}
    \vspace*{-0.1cm}
    \caption{Qualitative examples where our Bayesian Selective Fusion approach successfully localizes, while NetVLAD produces incorrect place matches, even when given the best reference set.}
    \label{fig:qualitative}
    \vspace*{-0.2cm}    
\end{figure}

Our approach can be seen to select informative reference sets (also shown in Fig.~\ref{fig:my_label}) since it attains equal or better performance than the \emph{single-reference} Bayesian methods. Such a scenario is shown in Fig.~\ref{fig:study_2}(b) on Nordland Winter, where our method's AUC is higher than the mean AUC of the single-reference Bayesian methods by $5\%$ and the best by $3\%$.
Finally, our approach outperforms NetVLAD on all but the easiest case (i.e., Oxford Sun with an absolute difference in AUC of less than $0.003$); improvements in AUC range from $8.5\%$ on Oxford Night to $25\%$ on the more difficult Nordland Winter.
Examples of correct matches from our approach and false matches from NetVLAD are shown in Fig.~\ref{fig:qualitative}.

\subsection{Comparison with Alternative Fusion Approaches}
\label{subsec:study_3}
We now perform an ablation-type study to examine the performance of our approach with and without Bayesian fusion and/or reference set selection. 
We consider a \emph{Bayesian Full Fusion} approach that omits reference selection; and, a \emph{Baseline Selective Fusion} approach based on the Baseline Fusion method of \eqref{eq:MED} that uses only the selected reference sets $\mathcal{S}$.
We also consider NetVLAD (advantageously) given the best reference on each traversal.

\begin{figure}[t!]
    \vspace{0.2cm} %
    \centering
    \includegraphics[width = \columnwidth]{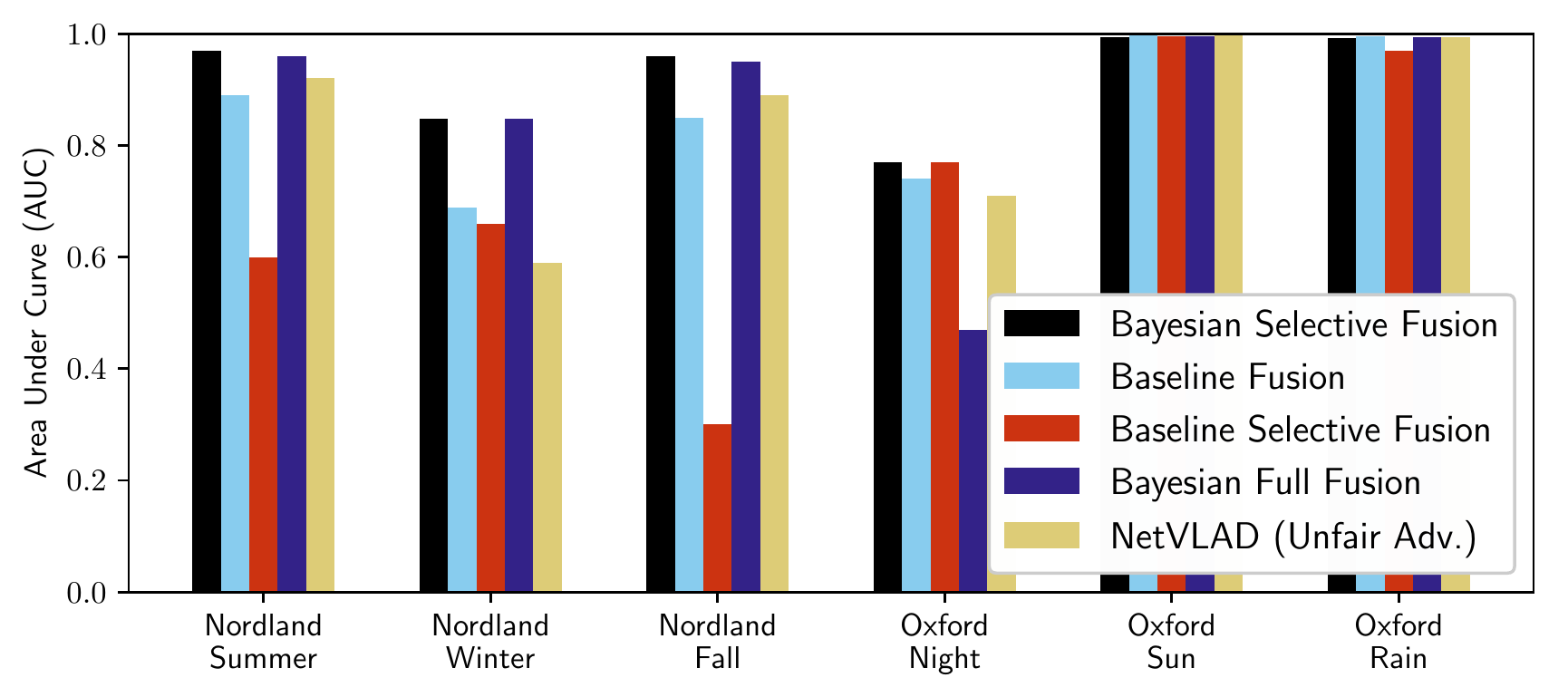}
     \caption{Area Under Precision-Recall Curves of Fig.~\ref{fig:study_3} for query images from six traversals from Nordland and Oxford RobotCar datasets.}
    \label{fig:study_3_auc}
    \vspace{-0.1cm}
 \end{figure}

\begin{figure}[t!]
    \vspace{0.2cm} %
    \captionsetup[subfigure]{aboveskip=-1.5pt,belowskip=0.5pt,labelformat=simple}
    \renewcommand*{\thesubfigure}{\quad(\alph{subfigure})}
    \centering
    \hspace{0.6cm}\includegraphics[width = 0.8\columnwidth]{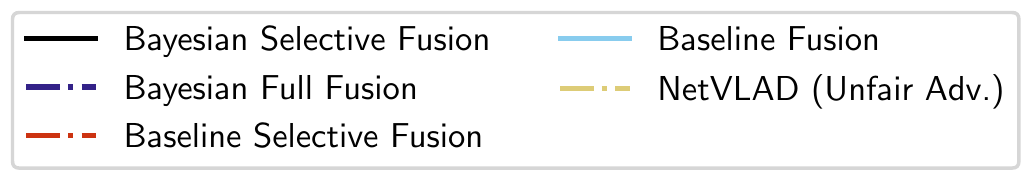}
 \begin{subfigure}{0.49\columnwidth}
     \centering
     \includegraphics[width = \textwidth]{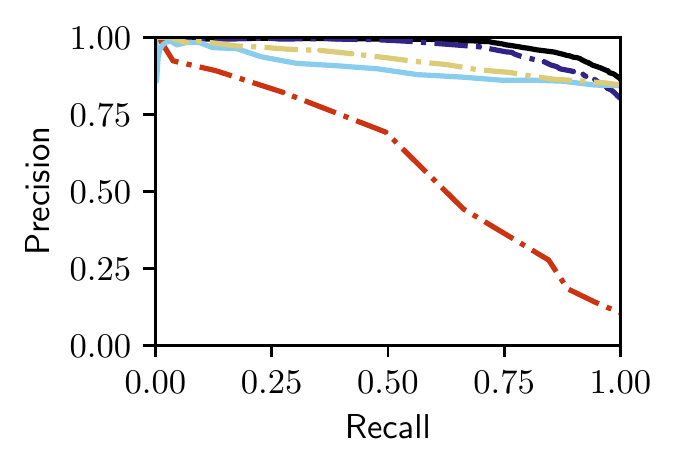}
     \caption{Query: Nordland Summer}
 \end{subfigure}
 \hfill
 \begin{subfigure}{0.49\columnwidth}
     \centering
     \includegraphics[width = \textwidth]{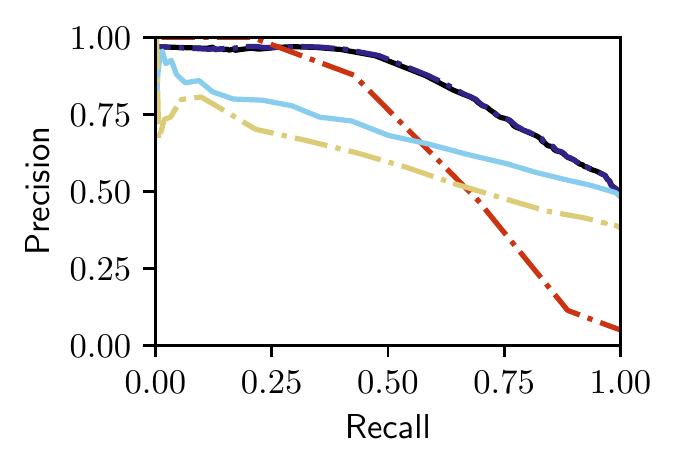}
     \caption{Query: Nordland Winter}
 \end{subfigure}\\
  \begin{subfigure}{0.49\columnwidth}
     \centering
    \includegraphics[width = \textwidth]{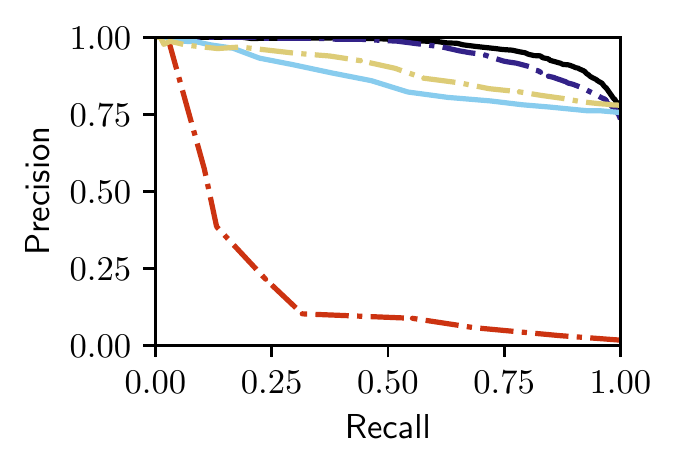}
     \caption{Query: Nordland Fall}
 \end{subfigure}
 \hfill
  \begin{subfigure}{0.49\columnwidth}
     \centering
        \includegraphics[width = \textwidth]{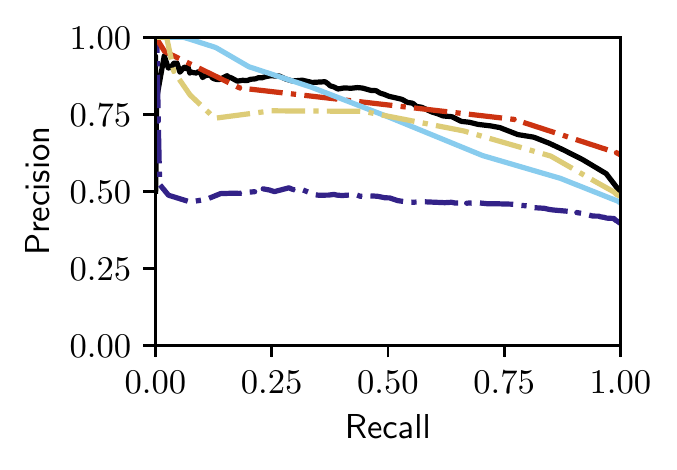}
     \caption{Query: Oxford Night}
 \end{subfigure}\\
 \begin{subfigure}{0.49\columnwidth}
     \centering 
        \includegraphics[width = \textwidth]{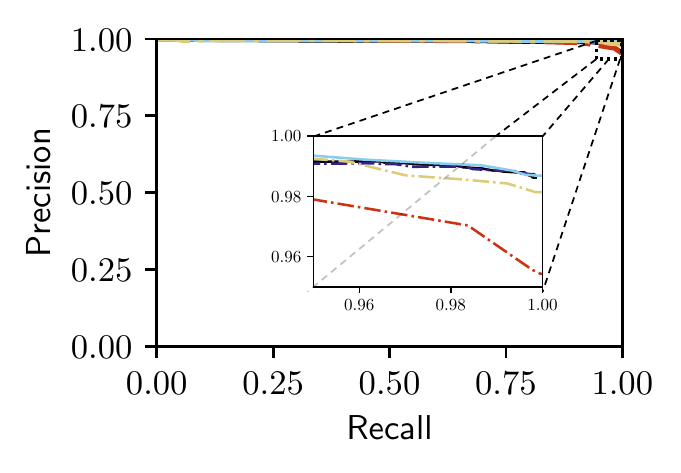}
     \caption{Query: Oxford Sun}
 \end{subfigure}
 \hfill
      \begin{subfigure}{0.49\columnwidth}
     \centering
        \includegraphics[width = \textwidth]{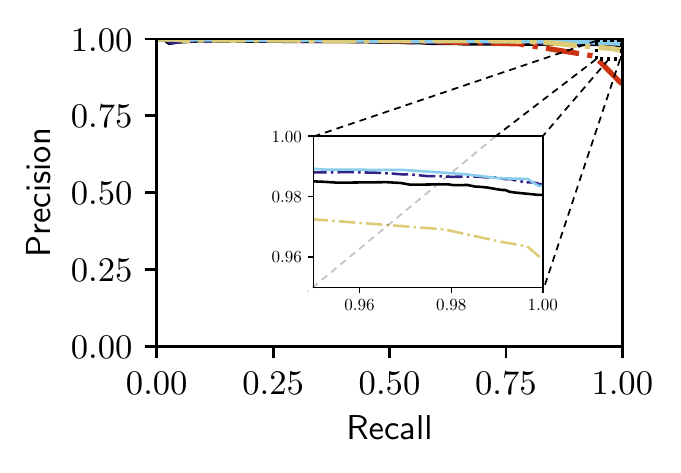}
     \caption{Query: Oxford Rain}
 \end{subfigure}
    \caption{Precision-recall comparison of Bayesian Selective Fusion approach with alternative fusion approaches on query images from six traversals from Nordland and Oxford RobotCar datasets.}
    \label{fig:study_3}
\end{figure}

Figs.~\ref{fig:study_3_auc}~and~\ref{fig:study_3} show that only our proposed Bayesian Selective Fusion approach works consistently well without any modification or parameter tuning across six query traversals from the Nordland and Oxford datasets.
All other approaches have instances of total failure or inferior performance.
For example, the Baseline Selective Fusion method fails almost completely on the difficult Nordland Summer, Winter, and Fall traversals, whilst the Baseline Fusion method and NetVLAD perform poorly.
The Bayesian Full Fusion approach compares the most favorably with our proposed Bayesian Selective Fusion approach but does not outperform it, illustrating the additional benefit of reference set selection.

\subsection{Extreme Appearance Changes}
\label{subsec:study_4}
Our reference set selection approach is motivated in part by the need for VPR systems to adapt in situations where the visual appearance of places can change dynamically and unpredictably such as indoors with a flickering light.
For this study, we consider a scenario in which the query image at each place in the Oxford dataset is drawn randomly (with equal probability) from either the Sun or Night traversals.

\begin{figure}[t!]
    \centering
    \includegraphics[width = 0.9\columnwidth]{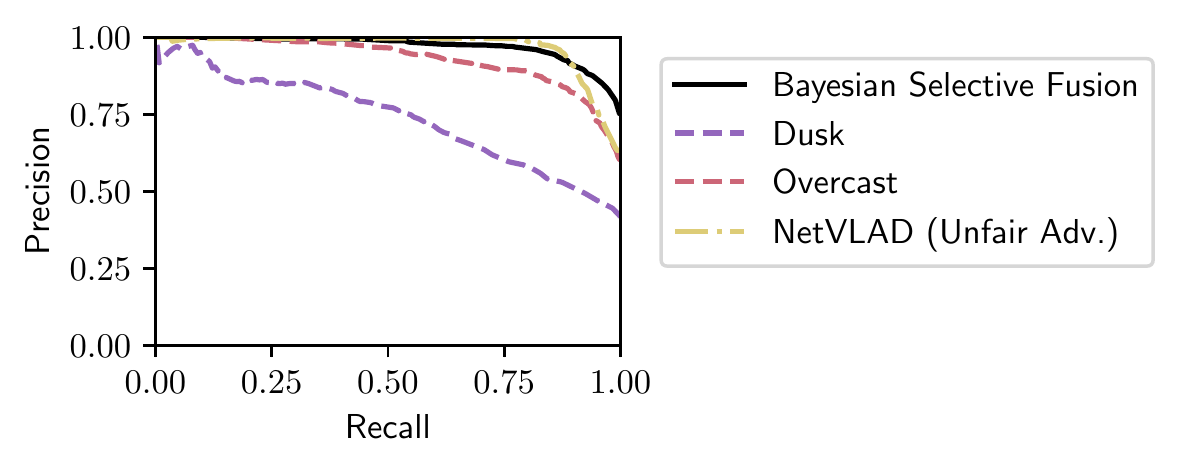}
    \vspace*{-0.3cm}
    \caption{Precision-recall on randomized query images from Sun and Night traversals from the Oxford RobotCar dataset.}
    \label{fig:study_4}
    \vspace*{-0.1cm}
\end{figure}

The performance of our proposed approach on the randomized Sun/Night query traversal is shown in Fig.~\ref{fig:study_4} together with the performance of the Dusk and Overcast single-reference Bayesian methods that perform best on the standalone Oxford Sun and Night traversals.
We also consider NetVLAD with it again given the advantage of the best reference set (overcast).
From Fig.~\ref{fig:study_4}, we see that Bayesian Selective Fusion outperforms the single-reference approaches and (the advantaged) NetVLAD with higher precision at recalls greater than $0.9$, and AUC improvements over (the advantaged) NetVLAD, Overcast, and Dusk of $1.0\%$, $6.6\%$, and $29\%$, respectively.\footnote{We expect worse performance from single-reference methods when query images can originate from more than two conditions (e.g.~Raining Night/Day, Snowing Night/Day, Raining Dusk/Dawn, etc.) but we lack sufficient data to perform these studies.}

\subsection{Sequence Matching}
\label{subsec:study_seq}
\begin{figure}[t!]
    \vspace{0.2cm} %
    \centering
    \includegraphics[width = 0.9\columnwidth]{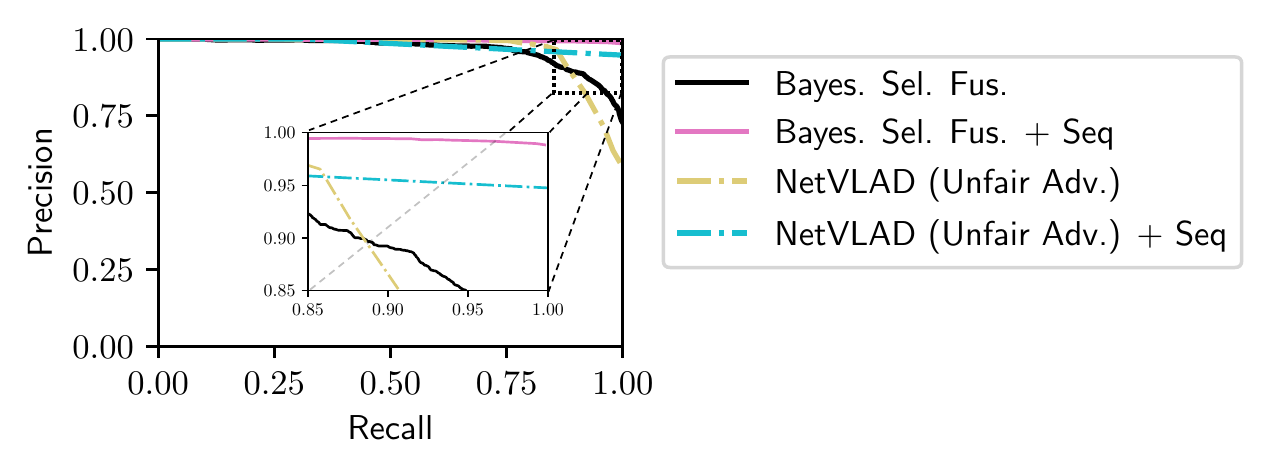}
    \vspace*{-0.2cm}
    \caption{Precision-recall of methods with and without sequence matching on randomized query images from Sun and Night traversals from the Oxford RobotCar dataset.}
    \label{fig:study_seq}
    \vspace*{-0.1cm}
\end{figure}

Sequence matching techniques are commonly used in conjunction with single-image place matching for VPR under challenging environments.
Fig.~\ref{fig:study_seq} shows that the performance of our approach on the randomized query from our previous study is improved via integration with an existing sequence matching approach (SeqSLAM \cite{Milford2012}) with a sequence length \mbox{of $5$}. 
It also performs better than NetVLAD integrated with the same sequence matching approach despite NetVLAD having prior knowledge of the best reference set.

\subsection{Other Image Descriptors}
\label{subsec:study_dense}
\begin{figure}[t!]
    \captionsetup[subfigure]{aboveskip=-1.5pt,belowskip=0.5pt,labelformat=simple}
    \renewcommand*{\thesubfigure}{\quad(\alph{subfigure})}
    \centering
    \includegraphics[width = 0.85\columnwidth]{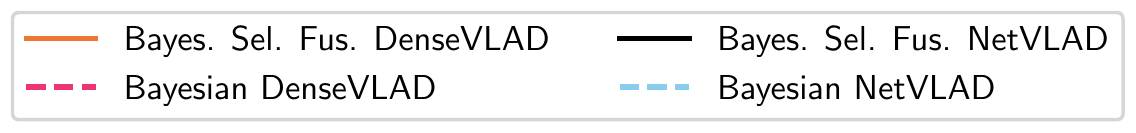}
 \begin{subfigure}{0.49\columnwidth}
     \centering
     \includegraphics[width = \textwidth]{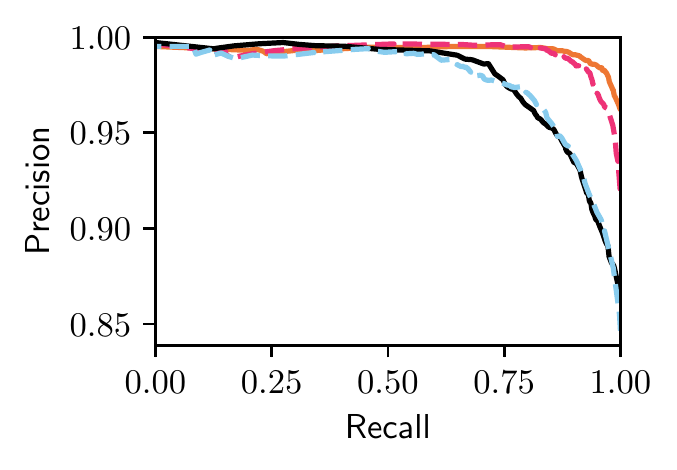}
     \caption{Nordland Summer}
 \end{subfigure}
 \hfill
 \begin{subfigure}{0.49\columnwidth}
     \centering
     \includegraphics[width = \textwidth]{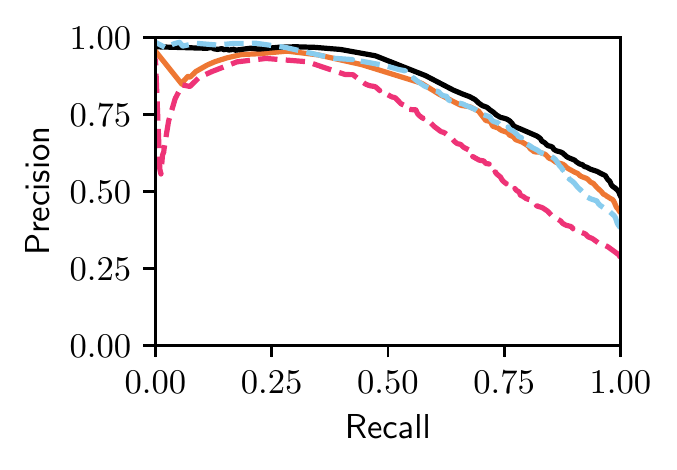}
     \caption{Nordland Winter}
 \end{subfigure}
    \caption{Precision-recall of Bayesian Selective Fusion and best single-reference Bayesian methods on query images from Nordland Summer and Winter traversals using DenseVLAD or NetVLAD image descriptors.}
    \label{fig:study_dense}
\end{figure}

In our previous studies, we used NetVLAD~\cite{NetVLAD} descriptors.
Fig.~\ref{fig:study_dense} shows the results of repeating our study from Section \ref{subsec:study_2} for Nordland summer and winter traversal queries with DenseVLAD~\cite{DenseVLAD} descriptors instead of NetVLAD descriptors.
We see that our approach delivers a similar performance gain compared to the best single-reference Bayesian methods without any modification, parameter tuning, or training.

\subsection{Compute Time and Scaling}
On an i7-8750H CPU, our approach took an average of $0.0224$s ($0.0002$s for reference selection and $0.0222$s for fusion) per query image with $M = 3$ and $N = 3000$ places.
Complexity is linear in the number of references $M$, however, the expensive Bayesian operations (\eqref{eq:approach} and \eqref{eq:obsLikelihoods}) scale linearly with the number of \emph{selected} reference sets in $\mathcal{S}$.
\section{Conclusions}
\label{sec:conclusion}
Visual place recognition relies on comparing query images with reference images previously captured and stored.
The storage and exploitation of multiple same-place reference images has garnered surprisingly little recent attention, with state-of-the-art approaches either attempting to determine the utility of reference images \emph{a priori} without consideration of the conditions at query time, or fusing all available reference images, which can be unnecessary and counterproductive.
In this paper, we have proposed Bayesian Selective Fusion as a novel approach to dynamically determine and exploit the utility of reference images at query time.
We have demonstrated that our approach can exceed the performance of state-of-the-art techniques on challenging benchmark datasets, including state-of-the-art single-reference methods provided with (advantageous) prior knowledge of the best reference images.
We have also shown that our approach is descriptor-agnostic and can be used in conjunction with standard techniques including sequence matching to achieve state-of-the-art performance.

Our current work can be extended in several ways, including the use of Bayesian Selective Fusion with recently proposed Delta Descriptors~\cite{DeltaDescriptors}, local descriptors, and probabilistic sequence matching via the prior in \eqref{eq:bayes}. 
We are also interested in evaluating our approach in situations with large viewpoint variations between the query and reference images but are currently limited by existing datasets offering too few reference sets. %
Furthermore, it would be interesting to explore the insight from our reference set selection approach to determine which extra reference sets should be collected, or which can be discarded to minimize storage requirements.
Finally, we believe that our research contributes to a better understanding of how to exploit the information from deep-learnt image descriptors within a training-free Bayesian setting, opening the possibility of using temporal Bayesian filtering in conjunction with reference set fusion, and further exploiting information-theoretic techniques in visual place recognition.

\bibliographystyle{IEEEtran}
\bibliography{IEEEabrv,Library,VPR}

\end{document}